\documentclass{article}

\PassOptionsToPackage{numbers,sort&compress}{natbib}

\usepackage[preprint]{neurips_2026}
\makeatletter
\def\@noticestring{}
\makeatother

\usepackage[utf8]{inputenc}
\usepackage[T1]{fontenc}

\usepackage{amsmath}
\usepackage{amssymb}
\usepackage{amsfonts}

\usepackage{booktabs}
\usepackage{tabularx}
\usepackage{array}
\usepackage{multirow}
\usepackage{fontawesome5}

\usepackage{xcolor,colortbl,multirow,booktabs,xfp}

\usepackage{enumitem}

\usepackage[table]{xcolor}
\definecolor{tablepurple}{HTML}{F2F2FF}
\definecolor{lightpurple}{RGB}{238,232,248}
\definecolor{mycomment}{HTML}{D62728}

\usepackage[most]{tcolorbox}
\usepackage{listings}

\definecolor{promptgray}{RGB}{248,248,248}
\definecolor{promptblue}{RGB}{245,248,255}
\definecolor{promptgreen}{RGB}{246,252,246}
\definecolor{promptred}{RGB}{255,246,246}

\lstdefinestyle{actionseq}{
  basicstyle=\ttfamily\footnotesize,
  breaklines=true,
  columns=fullflexible,
  frame=single,
  keepspaces=true
}

\lstdefinestyle{tgcode}{
  basicstyle=\ttfamily\footnotesize,
  breaklines=true,
  columns=fullflexible,
  frame=single,
  keepspaces=true
}

\usepackage{nicefrac}
\usepackage{url}

\usepackage{hyperref}
\usepackage{fontawesome5}

\definecolor{appendixblue}{RGB}{83,86,158}

\newcommand{\appgroup}[1]{%
    \vspace{0.75em}
    \noindent{\bfseries #1}\par
    \vspace{0.15em}
}

\newcommand{\appentrydesc}[3]{%
    \noindent
    \textcolor{appendixblue}{\textbf{Appendix~\ref{#1}: #2}}%
    \leaders\hbox to 0.55em{\hss.\hss}\hfill
    \textbf{\pageref{#1}}\par
    \noindent\hspace{1.6em}{\small #3}\par
    \vspace{0.45em}
}

\title{\textsc{TaskGround}: Structured Executable Task Inference for Full-Scene Household Reasoning}

\usepackage{marvosym} % for \Letter
\newcommand{\corrsym}{\raisebox{0.15ex}{\scriptsize\Letter}}
\usepackage{caption}
\usepackage{marvosym} % for \Letter

\newcommand{\internsym}{\ensuremath{\dagger}}
\author{%
\makebox[\textwidth][l]{%
\begin{tabular}{@{}l@{\quad}l@{\quad}l@{\quad}l@{\quad}l@{}}
\textbf{ZhiYuan Feng}\textsuperscript{1,\internsym} &
\textbf{Yu Deng}\textsuperscript{2} &
\textbf{Ruichuan An}\textsuperscript{3,\internsym} &
\textbf{Zhenhua Liu}\textsuperscript{2} &
\textbf{Qixiu Li}\textsuperscript{1,\internsym} \\
\textbf{Keming Wu}\textsuperscript{1} &
\textbf{Zhiying Du}\textsuperscript{4,\internsym} &
\textbf{Weijie Wang}\textsuperscript{5} &
\textbf{Haoxiao Wang}\textsuperscript{5} &
\textbf{Shuang Chen}\textsuperscript{5} \\
\textbf{Sicheng Xu}\textsuperscript{2} &
\textbf{Yaobo Liang}\textsuperscript{2} &
\textbf{Jiaolong Yang}\textsuperscript{2,\corrsym} &
\textbf{Baining Guo}\textsuperscript{2} &
\end{tabular}%
} \\
\vspace{0.45em}
\makebox[\textwidth][l]{%
\begin{tabular}{@{}l@{\qquad}l@{\qquad}l@{}}
\textsuperscript{1}Tsinghua University &
\textsuperscript{2}Microsoft Research Asia &
\textsuperscript{3}Peking University \\
\textsuperscript{4}Fudan University &
\textsuperscript{5}Zhejiang University &
\end{tabular}%
} \\
\vspace{0.55em}
\makebox[\textwidth][c]{%
\small
\href{https://aaronfengzy.github.io/TaskGround/}{\faHome\ Project Page}
\qquad
\href{https://github.com/AaronFengZY/TaskGround}{\faGithub\ Code}
}
}
\begin{document}

\maketitle

\begingroup
\renewcommand{\thefootnote}{\internsym}
\footnotetext{Work done during a research internship at Microsoft Research Asia.}
\renewcommand{\thefootnote}{\corrsym}
\footnotetext{Corresponding author.}
\endgroup
\setcounter{footnote}{0}
\vspace{-2.8em}
\begin{center}
\begin{minipage}{0.95\textwidth}
    \centering
    \captionsetup{type=figure,hypcap=false,skip=2pt}
    \includegraphics[width=\linewidth]{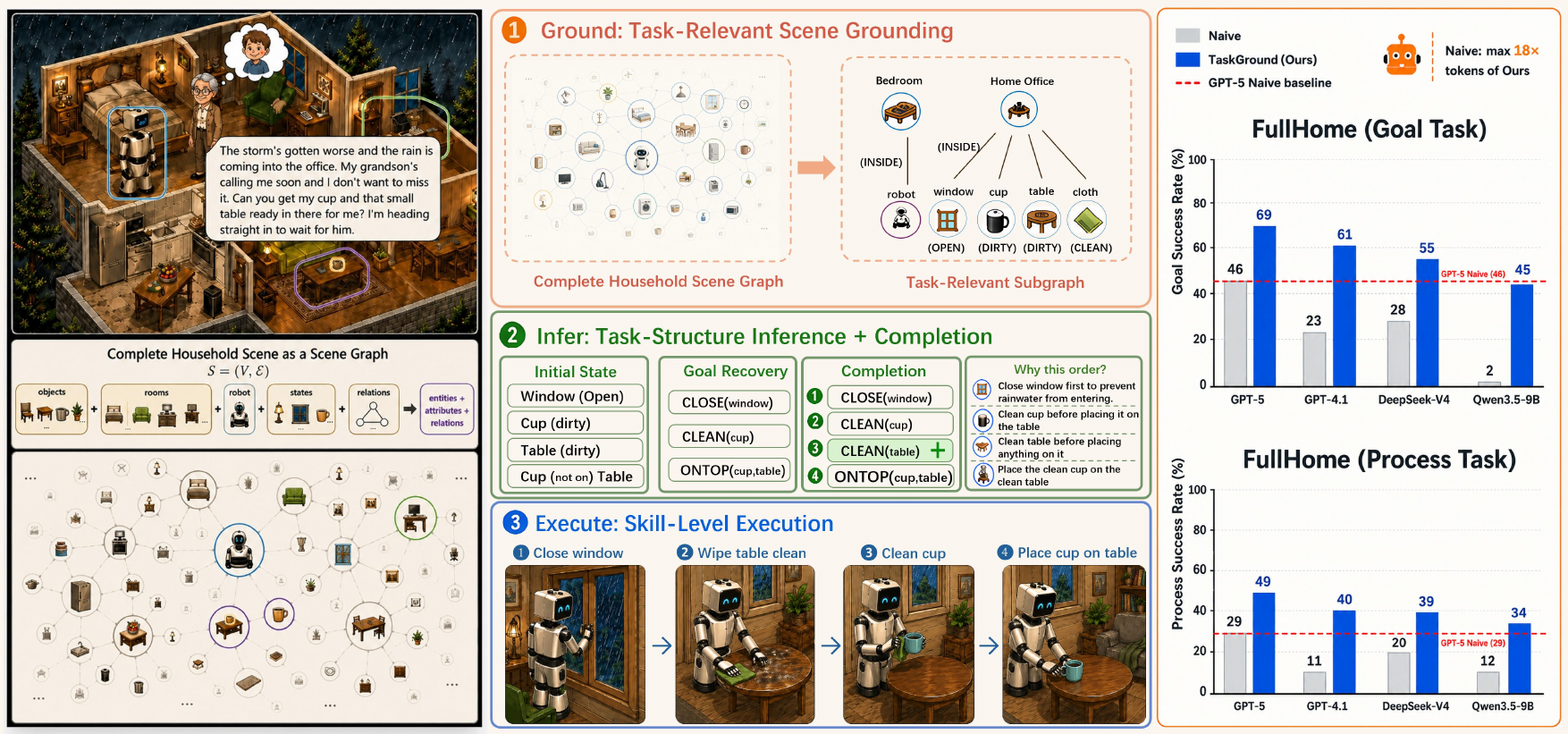}
    \captionof{figure}{
    Overview of \textbf{TaskGround} for full-scene household reasoning.
    It grounds complete scenes, infers executable task structure, and compiles grounded skill-level actions, improving task success rates on our FullHome benchmark with up to \(18\times\) lower total input-token cost.
    }
    \label{fig:teaser}
\end{minipage}
\end{center}

\vspace{0.8em}

\begin{abstract}
In real home deployments, household agents must often operate from a complete household scene and a situated household request, rather than from a clean task specification.
Such requests require agents to identify task-relevant entities, recover intended task conditions, and resolve ordering constraints from the surrounding scene context.
We formalize this capability as \textbf{full-scene household reasoning}: given a complete household scene and a situated household request, an agent must infer executable task structure before producing a grounded skill-level action sequence.
This setting is challenging because complete household scenes contain substantial task-irrelevant information, making direct complete-scene prompting inefficient and error-prone.
In practical deployment, this challenge is further amplified by privacy and local compute constraints, which favor compact open-weight models with limited long-context reasoning ability.
We propose \textbf{TaskGround}, a training-free and model-agnostic \emph{Ground--Infer--Execute} framework that grounds complete scenes into compact task-relevant scene slices, infers executable task structure, and compiles it into grounded skill-level action sequences.
To evaluate this setting, we introduce \textbf{FullHome}, a human-validated evaluation suite of 400 household tasks spanning diverse home-scale environments and both goal-oriented and process-constrained requirements.
On FullHome, TaskGround improves task success rates by large margins across both proprietary and open-weight models.
Notably, it makes Qwen3.5-9B competitive with GPT-5 under direct complete-scene prompting while reducing total input-token cost by up to 18$\times$.
Our results identify executable task-structure inference as a central bottleneck in full-scene household reasoning and show that structured grounding can make compact local models substantially more effective for practical household deployment.
\end{abstract}
\section{Introduction}
In real home deployments, household requests are often embedded in everyday situations rather than written as clean task specifications.
Consider the example in Fig.~\ref{fig:teaser}: a grandmother notices rain blowing into her home office while waiting for a call from her grandson.
She says to her household robot,
\emph{``The storm is getting worse, and rain is coming into the office.
My grandson is calling soon.
Can you get my cup and the small table ready in there?''}
Although natural, this request is far from a clean task specification.
It does not explicitly identify all task-relevant entities, such as the office window, the dirty cup, the table, and the target room.
Nor does it specify the intended task conditions or process requirements. Given the scene context, however, the robot should infer that it needs to close the window to prevent further rain from entering, clean the table before placing the cup on it, clean the dirty cup, and finally place the cup on the table.
Thus, before generating actions, the agent must infer an intermediate representation from the complete scene and the context-dependent request.
We refer to this representation as \emph{executable task structure}: the task-relevant entities, target conditions, relations, and process requirements sufficient for producing executable actions.
This example motivates \textbf{full-scene household reasoning}.
Given a \emph{complete household scene} and a \emph{situated household request}, an agent must infer executable task structure and produce a grounded skill-level action sequence.
Here, a complete household scene contains the full scene context available to the agent, including both task-relevant and irrelevant objects, states, and relations.
A situated household request is a naturally phrased, context-dependent request whose intended meaning is grounded in the current household scene rather than fully specified by language alone.
The output is a grounded skill-level action sequence, such as walking to the window, closing it, cleaning the table, cleaning the cup, and placing the cup on the table.

Recent progress in embodied planning with large language models (LLMs) has shown that LLMs can support high-level household decision making, including action-sequence and programmatic plan generation~\citep{llmzeroshot,progprompt,llmplanner,tapa,mldt}, skill or policy grounding~\citep{saycan,codeaspolicies}, and interactive or personalized assistance~\citep{innermonologue,tidybot}.
However, these systems typically study a different information condition from full-scene household reasoning.
They often assume that the user instruction, task goal, relevant objects, or executable constraints have already been stated, structured, or narrowed before planning begins.
As a result, the main challenge is often to map a given task description to executable actions or skills.
In contrast, our setting starts from a \emph{complete household scene} and a \emph{situated household request}, and requires the agent to produce a grounded skill-level action sequence.
To succeed, the agent must first infer \emph{executable task structure}, which specifies the relevant entities, target conditions, and process requirements needed to generate executable actions.
Thus, full-scene household reasoning evaluates whether an agent can infer what to plan for from complete scene context, rather than only whether it can execute a task whose structure has already been exposed.

This information condition is especially important for practical home deployment.
A complete household scene may contain many rooms, objects, states, affordances, and relations, making direct prompting over complete scenes costly and error-prone.
Detailed household states can also reveal private information about personal spaces, daily routines, and object usage patterns, making it undesirable to rely solely on frontier proprietary models for complete-scene inputs~\citep{denning2009spotlight,pagallo2013robots,bugeja2021prash,windl2024privacy}.
Meanwhile, real homes vary widely in layout, object distribution, routines, and user preferences, so practical systems should adapt to new households without task-specific training or repeated model tuning.
Local or on-device deployment further imposes memory, compute, energy, and latency constraints, favoring compact open-weight models that are easier to deploy but less capable of reasoning over long, cluttered inputs~\citep{qu2025mobile,wang2025empowering,zheng2025review}.
These challenges motivate a low-cost, model-agnostic mechanism for inferring executable task structure from complete household scenes and situated household requests under limited model capacity.

Motivated by this need, we propose \textbf{TaskGround}, a training-free and model-agnostic \emph{Ground--Infer--Execute} framework for full-scene household reasoning.
As illustrated in Fig.~\ref{fig:teaser}, TaskGround avoids directly generating a grounded skill-level action sequence from a long and cluttered complete household scene.
Instead, given a complete household scene and a situated household request, it first grounds the scene into a compact task-relevant scene slice by querying the language model over scene entities and reconstructing the needed local context around selected entities, substantially reducing the input cost of downstream reasoning.
Over this grounded context, TaskGround infers \emph{executable task structure}, specifying the task-relevant entities, target conditions, and ordering requirements needed for execution.
Because situated household requests often leave goals or intermediate steps implicit, TaskGround further completes the inferred structure with non-oracle household priors and process-critical subgoals, reflecting generic household patterns or pre-specified routines without using ground-truth task structures or evaluation outcomes.
Finally, TaskGround compiles the completed structure into a grounded skill-level action sequence through a task-agnostic skill interface, such as walking to an object, closing a window, cleaning a table, and placing a cup on it.
This decomposition reduces the reasoning burden on the base model while preserving executability, making TaskGround efficient, compatible with both proprietary and open-weight models, and suitable for deployment without task-specific training.

To evaluate this problem systematically, we introduce \textbf{FullHome}, a
simulator-backed evaluation suite of 400 carefully designed and human-validated household tasks.
Unlike many prior household environments and embodied-agent interfaces that often expose programmatic task specifications, explicit goals, curated objects, or structured constraints before planning~\citep{virtualhome,behavior,behavior1k,eai,partnr}, FullHome evaluates agents under the full-scene information condition: given a complete household scene and a situated household request, the agent must produce a grounded skill-level action sequence without being given task-relevant objects, target states, or process requirements.
We conduct comprehensive experiments on FullHome across proprietary and
open-weight models.
TaskGround consistently improves success rates across model families.
Notably, it makes Qwen3.5-9B competitive with GPT-5 under direct complete-scene prompting, while reducing total input-token cost by up to 18$\times$.
Ablation studies further confirm that scene grounding, task-structure
inference, task-structure completion, and skill-level execution each contribute to the final performance.
Together, these results demonstrate that structured grounding can make compact local models substantially more effective for full-scene household reasoning without task-specific training.

In summary, our main contributions are as follows:
\begin{itemize}
    \item We formalize \textbf{full-scene household reasoning} as a realistic information condition for household agents: given a complete household scene and a situated household request, the agent must produce a grounded skill-level action sequence without being given task-relevant objects, target states, or process requirements.

    \item We propose \textbf{TaskGround}, a training-free and model-agnostic \emph{Ground--Infer--Execute} framework for this setting. TaskGround grounds complete household scenes into compact task-relevant scene slices, infers and completes executable task structure, and compiles the completed structure into grounded skill-level action sequences.

    \item We introduce \textbf{FullHome}, a simulator-backed evaluation suite for systematically measuring full-scene household reasoning. FullHome contains 400 carefully designed and human-validated household tasks across diverse home-scale environments, covering both goal-oriented and process-constrained requirements.

    \item We conduct comprehensive experiments across proprietary and open-weight models. TaskGround consistently improves success rates, makes Qwen3.5-9B competitive with GPT-5 under direct complete-scene prompting, and reduces total input-token cost by up to 18$\times$ on larger complete scenes.
\end{itemize}
\section{Problem Formulation}
\label{sec:problem}
We formalize \textbf{full-scene household reasoning} as the task of mapping a
\emph{complete household scene} and a \emph{situated household request} to a
grounded skill-level action sequence.
A complete household scene is represented as a scene graph
\(\mathcal{S}=(\mathcal{V},\mathcal{E})\), where nodes in \(\mathcal{V}\)
denote household entities, such as rooms, objects, surfaces, containers and the
agent, and edges in \(\mathcal{E}\) denote relations among them, such as
containment, support, proximity, and room membership.
Each entity may also carry attributes describing its category, current state,
and available affordances.
The scene is \emph{complete} in the sense that it contains the full household
context available to the agent, including both task-relevant and irrelevant
entities.

An instance of full-scene household reasoning consists of an initial complete
household scene \(\mathcal{S}_0\) and a situated household request \(u\).
A situated household request is a naturally phrased, context-dependent user
request whose intended meaning is grounded in the current household scene rather
than fully specified by language alone.
The agent takes \((\mathcal{S}_0,u)\) as input and outputs a grounded
skill-level action sequence:
\[
    (\mathcal{S}_0,u)
    \xrightarrow{\;\text{agent}\;}
    a_{1:T}=(a_1,a_2,\ldots,a_T).
\]
Each action \(a_t\) is an executable skill-level command with grounded entity
arguments, such as \texttt{WalkTo(window)}, \texttt{Close(window)},
\texttt{Wipe(table)}, \texttt{Clean(cup)}, or \texttt{Place(cup, table)}.
These skills abstract low-level control into object-centric actions that a
household agent is assumed to be able to perform, and each action must refer to
entities in the current scene.
Executing a valid action updates the scene state, written as
\(\mathcal{S}_t=\operatorname{Step}(\mathcal{S}_{t-1},a_t)\), where
\(\operatorname{Step}\) denotes the environment transition induced by the
executed action.
Invalid actions, such as placing an object without holding it or putting an
object into a closed container, lead to execution failure.

Although the required output is only \(a_{1:T}\), the input pair
\((\mathcal{S}_0,u)\) implicitly defines what the agent should accomplish.
We describe this hidden task semantics as
\[
    \tau = (\mathcal{G}, \mathcal{P}),
\]
where \(\mathcal{G}\) is a set of goal conditions and \(\mathcal{P}\) is a set
of process constraints.
Each goal condition \(g\in\mathcal{G}\) describes a desired state or relation
that should hold after task completion.
For example, in the scenario in Fig.~\ref{fig:teaser}, the request may imply
goals such as \texttt{CLOSED(window)}, \texttt{CLEAN(table)},
\texttt{CLEAN(cup)}, and \texttt{ON(cup, table)}.
These goals specify the intended outcome, even though the user does not
explicitly list them in the request.

Some household requests also imply temporal or process requirements.
We represent these requirements with constraints in \(\mathcal{P}\), written in
the form \(g_i~\texttt{then}~g_j\), meaning that condition \(g_i\) should be
achieved before condition \(g_j\).
For example, the agent should close the window before continuing to prepare the
office area, and it should clean the table and cup before placing the cup on the
table.
This can be expressed with constraints such as
\(\texttt{CLOSED(window)}~\texttt{then}~\texttt{CLEAN(table)}\),
\(\texttt{CLEAN(table)}~\texttt{then}~\texttt{ON(cup, table)}\), and
\(\texttt{CLEAN(cup)}~\texttt{then}~\texttt{ON(cup, table)}\).
Multiple constraints can be combined by conjunction.
Such process constraints capture requirements that are not determined by the
final state alone.

Importantly, \(\tau=(\mathcal{G},\mathcal{P})\) is not given to the agent and
is not required as the output.
The agent only receives the complete household scene and the situated household
request, and must produce an executable grounded skill-level action sequence.
Full-scene household reasoning therefore tests whether an agent can infer the
task-relevant entities, intended goals, and process requirements from complete
scene context and situated language, rather than merely execute a task whose
objects, goals, and constraints have already been specified.
\section{TaskGround: Ground--Infer--Execute Framework}
\label{sec:framework}

The formulation above highlights three coupled challenges for deployable full-scene household reasoning.
First, the complete household scene can contain many irrelevant entities and relations, making direct complete-scene prompting costly and error-prone.
Second, the intended goals and process requirements are often implicit in the situated household request, so the agent must infer what task structure should be achieved before acting.
Third, the final output must be executable: actions must be grounded to available entities and satisfy skill preconditions, otherwise the execution may fail even when the high-level intent is correct.
Practical deployment further requires this capability without task-specific training whenever the home, user, or base model changes.
To address these challenges, we propose \textbf{TaskGround}, a training-free and model-agnostic \emph{Ground--Infer--Execute} framework.

\begin{figure}[t]
    \centering
    \includegraphics[width=\textwidth]{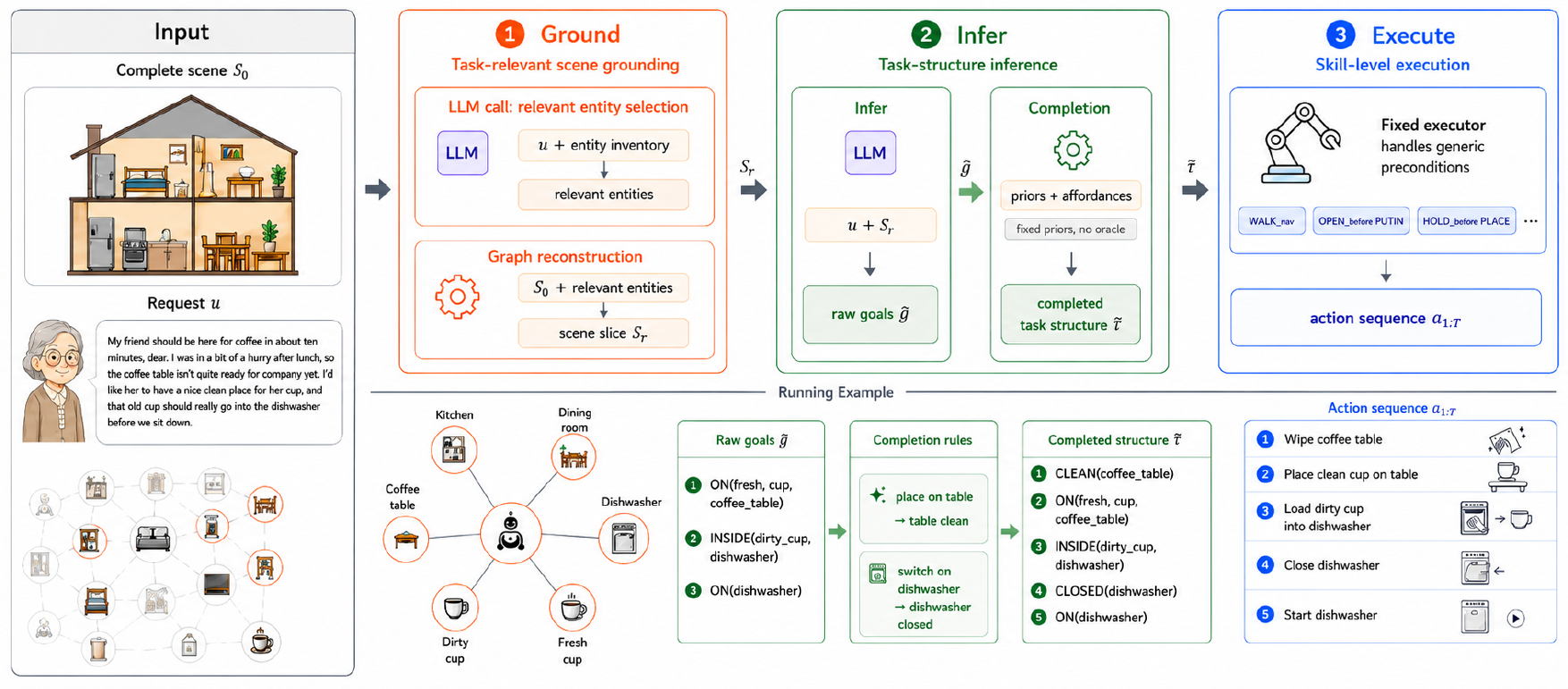}
    \caption{
    Overview of \textbf{TaskGround}.
    Given a complete household scene and a situated household request, TaskGround follows a \emph{Ground--Infer--Execute} decomposition.
    The orange module grounds the input into a compact task-relevant scene slice; the green module infers and completes executable task structure; the blue module compiles the completed structure into a grounded skill-level action sequence.
    The bottom row shows a running example from scene grounding to completed task structure and executable actions.
    }
    \label{fig:framework}
\end{figure}

As illustrated in Fig.~\ref{fig:framework}, TaskGround aligns these challenges with three stages.
Given an initial complete household scene \(\mathcal{S}_0\) and a situated household request \(u\), it first grounds the scene into a compact task-relevant scene slice, then infers and completes executable task structure, and finally compiles the completed structure into grounded skill-level actions:
\begin{equation}
\text{Ground: } \mathcal{S}_r = R(\mathcal{S}_0,u), \quad
\text{infer: } \hat{\mathbf{g}} = T(\mathcal{S}_r,u),\ \tilde{\tau}=C(\hat{\mathbf{g}}), \quad
\text{Execute: } a_{1:T}=E(\tilde{\tau},\mathcal{S}_r).
\label{eq:taskground}
\end{equation}
Here, \(R\), \(T\), \(C\), and \(E\) denote the scene grounder, task-structure inference module, completion module, and executor, respectively.
The completed structure \(\tilde{\tau}\) is an ordered goal sequence used for execution, and TaskGround does not access the hidden task semantics \(\tau=(\mathcal{G},\mathcal{P})\) defined in Sec.~\ref{sec:problem}.

\noindent\textbf{Ground: Task-Relevant Scene Grounding.}
The grounding stage produces a compact task-relevant scene slice
\(\mathcal{S}_r\) from the complete household scene \(\mathcal{S}_0\).
TaskGround first extracts an entity inventory from \(\mathcal{S}_0\), including
object and room names with lightweight state descriptions, and provides this
inventory together with the situated request \(u\) to the language model.
The model returns a set of task-relevant entities \(\hat{\mathcal{V}}_r\), which
may be explicitly mentioned in the request or implicitly required by the task.

TaskGround then reconstructs the grounded scene slice by expanding
\(\hat{\mathcal{V}}_r\) over the original scene graph.
Specifically, it adds the containing rooms, the agent node and state, and
execution-relevant neighboring entities when needed, and then restores all
relations in \(\mathcal{S}_0\) whose endpoints are included in the expanded
node set.
Formally, if \(\mathcal{V}_r\) is the expanded node set, the grounded slice is
\[
    \mathcal{S}_r=(\mathcal{V}_r,\mathcal{E}_r), \qquad
    \mathcal{E}_r=\{(v_i,v_j)\in\mathcal{E}_0 \mid v_i,v_j\in\mathcal{V}_r\}.
\]
This node-first grounding removes irrelevant clutter while preserving the local context needed for downstream task-structure inference and execution.
As reported in Sec.~\ref{sec:experiments}, this reduces total input-token cost by up to 18$\times$ compared with direct complete-scene prompting.

\noindent\textbf{Infer: Task-Structure Inference and Completion.}
Given the grounded scene slice \(\mathcal{S}_r\) and request \(u\), the inference
module \(T\) predicts an ordered goal sequence
\(\hat{\mathbf{g}}=(\hat{g}_1,\ldots,\hat{g}_M)\).
Each goal atom describes a desired state or relation over grounded entities,
such as \texttt{ON(fresh\_cup, coffee\_table)} or
\texttt{INSIDE(dirty\_cup, dishwasher)}.
This goal-level output focuses the language model on task interpretation before
skill-level action generation.

The initial sequence may still miss process-critical goals that are implicit in
household execution.
For example, as shown in Fig.~\ref{fig:framework}, the model may predict
\texttt{INSIDE(dirty\_cup, dishwasher)} and \texttt{ON(dishwasher)}, but omit
\texttt{CLOSED(dishwasher)}, which is required before starting the dishwasher.
The completion module \(C\) therefore augments \(\hat{\mathbf{g}}\) using fixed
household priors and object affordances, such as closing an appliance before
turning it on or cleaning a surface before placing objects on it.
These priors are specified before evaluation and do not access hidden goals,
oracle actions, task identifiers, or evaluation outcomes.
The result is a completed executable task structure
\(\tilde{\tau}=(\tilde{g}_1,\ldots,\tilde{g}_K)\), an ordered goal sequence that
contains both final goals and process-critical intermediate goals.

\noindent\textbf{Execute: Skill-Level Execution.}
The execution stage converts the completed executable task structure
\(\tilde{\tau}\) into a grounded skill-level action sequence \(a_{1:T}\).
Given \(\tilde{\tau}\) and the grounded scene slice \(\mathcal{S}_r\), the executor
\(E\) realizes each ordered goal atom as a scene-state change using the agent's
available skills, such as navigation, manipulation, cleaning, container
operation, or appliance control.
In Fig.~\ref{fig:framework}, this corresponds to the blue-panel action sequence.
The executor also handles generic preconditions required for executable actions,
such as navigating near an object before manipulation, holding an object before
placing it, or opening a container before insertion.
It does not infer task-relevant entities, implicit goals, or process
requirements; it only realizes the completed structure produced by the previous
stages.
This delegates action realization to a fixed skill-level executor, reducing
unsupported or hallucinated actions while preserving executability.
\section{FullHome Evaluation Protocol}
\label{sec:fullhome}

We instantiate full-scene household reasoning with \textbf{FullHome}, a
simulator-backed evaluation suite of 400 carefully designed and human-validated
household tasks built on executable household environments, including
VirtualHome~\citep{virtualhome} and BEHAVIOR~\citep{behavior}.
FullHome changes the test-time information condition of existing
simulator-backed household evaluations: the agent receives only the complete
household scene \(\mathcal{S}_0\) and the situated household request \(u\), and
must output a grounded skill-level action sequence \(a_{1:T}\).
It is not given task-relevant objects, candidate action arguments, explicit
goals, target states, reference action sequences, or process constraints.
The predicted sequence is executed in the simulator, producing a trajectory
\(\rho=(\mathcal{S}_0,a_1,\mathcal{S}_1,\ldots,a_T,\mathcal{S}_T)\), which is
evaluated against the hidden task structure \(\tau=(\mathcal{G},\mathcal{P})\).

FullHome contains two complementary task families.
We use \(\models\) to denote satisfaction.
For goal-oriented tasks, success requires the final scene state to satisfy all
goal conditions, i.e., \(\mathcal{S}_T \models \mathcal{G}\).
For process-constrained tasks, success additionally requires the execution
trajectory to satisfy the process constraints, i.e.,
\(\mathcal{S}_T \models \mathcal{G}\) and \(\rho \models \mathcal{P}\).
This protocol separates failures in reaching the desired final state from
failures in satisfying required intermediate steps or ordering constraints.
It therefore evaluates whether agents can infer executable task structure from
complete household scenes and situated household requests, rather than relying
on task structure exposed before planning. Additional benchmark construction and validation details are provided in
Appendix~\ref{app:benchmark_details}.
\section{Experiments}
\label{sec:experiments}

We study whether TaskGround improves full-scene household reasoning, which
components drive the gains, and whether structured grounding helps compact
open-weight models approach frontier direct prompting with lower input cost.

\subsection{Experimental Setup}
\label{sec:exp_setup}

FullHome contains 400 tasks. For each task, the agent receives only
\((\mathcal{S}_0,u)\) and must output a grounded skill-level action sequence,
without access to explicit task-relevant objects, goals, constraints, or
reference action sequences. We evaluate proprietary and open-weight models,
including GPT-4o~\citep{gpt4o}, GPT-4.1~\citep{gpt41},
Gemini-2.5-Flash~\citep{gemini25flash}, GPT-5~\citep{gpt5},
DeepSeek-V4-Flash~\citep{deepseekv4}, MiMo-V2-Flash~\citep{mimoflash},
Gemma-3-12B-it~\citep{gemma3}, and Qwen3.5-9B~\citep{qwen35}, covering both
frontier proprietary systems and compact models suitable for local deployment.
We report \textbf{Goal SR} and \textbf{Process SR}: Goal SR measures whether
the final executed state satisfies the goal conditions, while Process SR
additionally requires the execution trajectory to satisfy process constraints.
% Additional diagnostics are reported in Appendix~\ref{app:diagnostics}.

\subsection{Main Results}
\label{sec:main_results}

We compare \textbf{Naive} and \textbf{TaskGround}. Naive directly prompts the model with the complete household scene and situated household request to generate grounded skill-level actions. Table~\ref{tab:main_results} reports the results. TaskGround consistently improves over Naive across all models, environments, and task types, showing that direct complete-scene action generation is brittle under situated household requests, while structured grounding and executable task-structure inference provide robust gains. TaskGround also narrows the gap between compact open-weight models and frontier direct prompting: Qwen3.5-9B with TaskGround becomes competitive with GPT-5 under Naive prompting, matching or exceeding it in three of the four environment--metric comparisons. Moreover, TaskGround improves GPT-5 itself, suggesting that the gains come from a better problem decomposition rather than only compensating for weaker base models.

% =============================================================
%  Main results table — pink gradient relative to GPT-5 Naive
%  (Original two-column layout: Model | Method, with \cmidrule
%  separators between every model.)
%
%  Self-contained: needs xcolor / colortbl / multirow / booktabs / xfp
%  in the document preamble. Everything else is defined here.
% =============================================================

% --- colours -------------------------------------------------
\providecolor{tablepurple}{HTML}{E8E0F1}
\providecolor{pinkL}{HTML}{FCE4EC}        % > GPT-5 Naive baseline

% small coloured square for legend use
\providecommand{\tcsq}[1]{{\fboxsep=0pt\fcolorbox{#1}{#1}{\rule{0pt}{0.7em}\rule{0.7em}{0pt}}}}

% red delta annotation, e.g. \dlt{+41.0} -> (+41.0) in small red
\providecolor{deltared}{HTML}{C0392B}
\providecommand{\dlt}[1]{{\scriptsize\textcolor{deltared}{(#1)}}}

% --- core macro: colour cell if value > baseline -------------
% \hcd{<baseline>}{<value>}
\providecommand{\hcd}[2]{%
  \ifnum\fpeval{(#2)>(#1) ? 1 : 0}=1
    \cellcolor{pinkL}#2%
  \else
    #2%
  \fi
}
% bolded variant (best-in-class)
\providecommand{\hcdb}[2]{%
  \ifnum\fpeval{(#2)>(#1) ? 1 : 0}=1
    \cellcolor{pinkL}\textbf{#2}%
  \else
    \textbf{#2}%
  \fi
}

% --- column-specific shortcuts (baselines = GPT-5 Naive) -----
\providecommand{\hcVG}[1]{\hcd{45.5}{#1}}   % VirtualHome  Goal SR
\providecommand{\hcVP}[1]{\hcd{26.0}{#1}}   % VirtualHome  Process SR
\providecommand{\hcBG}[1]{\hcd{46.2}{#1}}   % BEHAVIOR     Goal SR
\providecommand{\hcBP}[1]{\hcd{31.4}{#1}}   % BEHAVIOR     Process SR
\providecommand{\hcVGb}[1]{\hcdb{45.5}{#1}}
\providecommand{\hcVPb}[1]{\hcdb{26.0}{#1}}
\providecommand{\hcBGb}[1]{\hcdb{46.2}{#1}}
\providecommand{\hcBPb}[1]{\hcdb{31.4}{#1}}

% =============================================================

\begin{table}[t]
\centering
\footnotesize
\setlength{\tabcolsep}{4.5pt}
\renewcommand{\arraystretch}{1.05}
\caption{%
Main results on FullHome.
\textbf{Goal SR} and \textbf{Process SR} are success rates on goal and process tasks.
Cells shaded \tcsq{pinkL} exceed \emph{GPT-5 Naive}, the frontier direct-prompting reference for each column.
\dlt{+$\Delta$} denote absolute gains of \emph{TaskGround} over the same model's \emph{Naive} result.
\textbf{Bold} marks the best result within each model group.
}
\label{tab:main_results}
% needs \usepackage{array} in the preamble (for the >{...} column spec)
\begin{tabular}{l p{1.8cm} cccc}
\toprule
\multicolumn{1}{c}{\multirow{2}{*}{\textbf{Model}}}
& \multicolumn{1}{c}{\multirow{2}{*}{\textbf{Method}}}
& \multicolumn{2}{c}{\textbf{VirtualHome}}
& \multicolumn{2}{c}{\textbf{BEHAVIOR}} \\
\cmidrule(lr){3-4} \cmidrule(lr){5-6}
& & \textbf{Goal SR} & \textbf{Process SR} & \textbf{Goal SR} & \textbf{Process SR} \\
\midrule

\rowcolor{tablepurple}
\multicolumn{6}{l}{\textbf{Proprietary models}} \\
\cmidrule(l{6pt}r{6pt}){1-6}

\multirow{2}{*}{GPT-4o}
& Naive      & \hcVG{8.0}  & \hcVP{5.0}  & \hcBG{32.3} & \hcBP{23.3} \\
& TaskGround & \hcVG{49.0}\,\dlt{+41.0} & \hcVP{37.0}\,\dlt{+32.0} & \hcBG{52.3}\,\dlt{+20.0} & \hcBP{35.5}\,\dlt{+12.2} \\
\cmidrule(l{6pt}r{6pt}){1-6}

\multirow{2}{*}{GPT-4.1}
& Naive      & \hcVG{14.0} & \hcVP{4.0}  & \hcBG{32.3} & \hcBP{17.1} \\
& TaskGround & \hcVG{57.0}\,\dlt{+43.0} & \hcVP{34.0}\,\dlt{+30.0} & \hcBGb{64.6}\,\dlt{+32.3} & \hcBP{45.7}\,\dlt{+28.6} \\
\cmidrule(l{6pt}r{6pt}){1-6}

\multirow{2}{*}{Gemini-2.5-Flash}
& Naive      & \hcVG{0.5}  & \hcVP{2.0}  & \hcBG{9.2}  & \hcBP{5.7}  \\
& TaskGround & \hcVG{48.5}\,\dlt{+48.0} & \hcVP{19.0}\,\dlt{+17.0} & \hcBG{38.5}\,\dlt{+29.3} & \hcBP{37.1}\,\dlt{+31.4} \\
\cmidrule(l{6pt}r{6pt}){1-6}

\multirow{2}{*}{GPT-5}
& Naive      & 45.5 & 26.0 & 46.2 & 31.4 \\
& TaskGround & \hcVGb{73.5}\,\dlt{+28.0} & \hcVPb{46.0}\,\dlt{+20.0} & \hcBGb{64.6}\,\dlt{+18.4} & \hcBPb{51.4}\,\dlt{+20.0} \\

\midrule
\rowcolor{tablepurple}
\multicolumn{6}{l}{\textbf{Open-weight models}} \\
\cmidrule(l{6pt}r{6pt}){1-6}

\multirow{2}{*}{DeepSeek-V4-Flash}
& Naive      & \hcVG{28.5} & \hcVP{15.0} & \hcBG{27.7} & \hcBP{25.7} \\
& TaskGround & \hcVGb{63.0}\,\dlt{+34.5} & \hcVPb{36.0}\,\dlt{+21.0} & \hcBGb{47.7}\,\dlt{+20.0} & \hcBPb{42.9}\,\dlt{+17.2} \\
\cmidrule(l{6pt}r{6pt}){1-6}

\multirow{2}{*}{MiMo-V2-Flash}
& Naive      & \hcVG{5.0}  & \hcVP{1.0}  & \hcBG{10.8} & \hcBP{11.4} \\
& TaskGround & \hcVG{43.5}\,\dlt{+38.5} & \hcVP{23.0}\,\dlt{+22.0} & \hcBG{43.1}\,\dlt{+32.3} & \hcBP{40.0}\,\dlt{+28.6} \\
\cmidrule(l{6pt}r{6pt}){1-6}

\multirow{2}{*}{Gemma-3-12B}
& Naive      & \hcVG{0.0}  & \hcVP{1.0}  & \hcBG{0.0}  & \hcBP{5.7}  \\
& TaskGround & \hcVG{37.0}\,\dlt{+37.0} & \hcVP{19.0}\,\dlt{+18.0} & \hcBG{29.2}\,\dlt{+29.2} & \hcBP{34.3}\,\dlt{+28.6} \\
\cmidrule(l{6pt}r{6pt}){1-6}

\multirow{2}{*}{Qwen3.5-9B}
& Naive      & \hcVG{0.0}  & \hcVP{10.0} & \hcBG{1.5}  & \hcBP{14.3} \\
& TaskGround & \hcVG{47.5}\,\dlt{+47.5} & \hcVP{29.0}\,\dlt{+19.0} & \hcBG{43.1}\,\dlt{+41.6} & \hcBP{40.0}\,\dlt{+25.7} \\

\bottomrule
\end{tabular}
\vspace{-0.6em}
\end{table}

\begin{figure}[b]
    \centering
    \includegraphics[width=\textwidth]{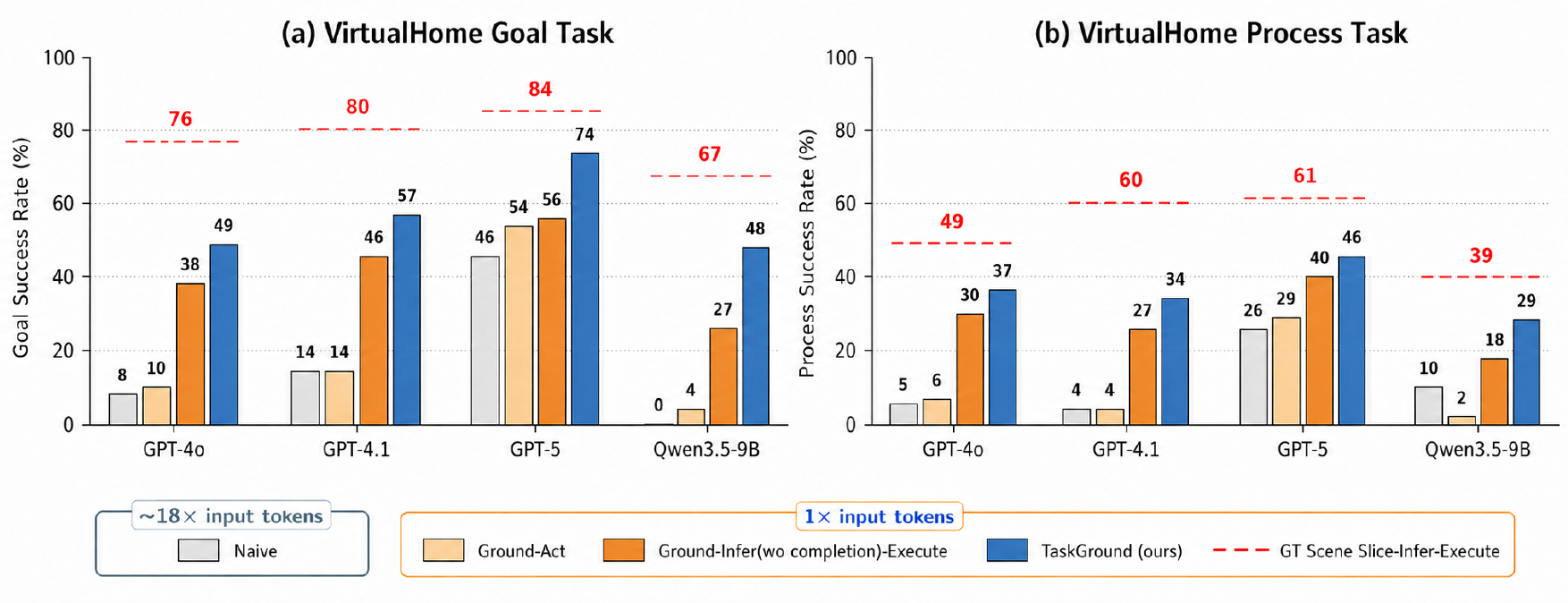}
    \caption{
    Framework ablation on the VirtualHome split of FullHome.
    Bars report \textbf{Goal SR} on goal-oriented tasks and \textbf{Process SR} on process-constrained tasks.
    The red dashed line shows a diagnostic GT-scene-slice setting that replaces the predicted scene slice with the task-relevant scene slice to isolate the grounding bottleneck.
    }
    \label{fig:framework_ablation}
\end{figure}

\subsection{Framework Ablation}
\label{sec:framework_ablation}

We conduct framework ablations on the VirtualHome split of FullHome, as shown in
Fig.~\ref{fig:framework_ablation}.
VirtualHome uses a single underlying household environment while varying the task-specific scene states, making it suitable for controlled component analysis and closer to a known-home deployment setting, where house-specific priors or routines can be specified before evaluation.
We compare five variants.
\textbf{Naive} directly generates skill-level actions from the complete scene and
request.
\textbf{Ground-Act} first applies scene grounding, but still asks the model to
directly generate actions from the grounded scene slice.
\textbf{Ground-Infer-Execute w/o Completion} removes the completion module from
TaskGround.
\textbf{TaskGround} uses the full framework.
Finally, \textbf{GT Scene Slice-Infer-Execute} replaces the predicted scene slice
with a task-relevant scene slice to estimate how much performance can improve if
grounding is near-perfect.
The variants also differ in input cost: \textbf{Naive} serializes the complete
household scene and uses about \(18\times\) total input tokens, while the
grounded-slice variants operate on compact task-relevant scene slices and are
normalized to \(1\times\) in Fig.~\ref{fig:framework_ablation}.

The ablation shows that each component contributes to the final performance.
Grounding reduces irrelevant scene clutter, but Ground-Act remains weak for
several models, indicating that a smaller input alone does not solve the task.
Adding task-structure inference gives large gains, especially for GPT-4o,
GPT-4.1, and Qwen3.5-9B, suggesting that these models struggle to directly turn a
scene and request into executable skill-level actions.
Completion further improves performance by recovering missing goals and
process-critical steps, such as closing an appliance before starting it or
cleaning a surface before placing objects on it.
The GT-scene-slice diagnostic remains above TaskGround, showing that scene
grounding is still a remaining bottleneck; meanwhile, its performance is still
not perfect, confirming that executable task-structure inference remains
challenging even with the relevant scene context.

% =============================================================
%  Grounding & task-structure recovery diagnostics
%  (VirtualHome split of FullHome, GPT-5 / Qwen3.5-9B,
%   4 methods x 2 task types x {Recall, Precision, Task Rec.}).
%
%  Self-contained: needs xcolor / colortbl / multirow / booktabs
%  in the document preamble.
%
%  Convention:
%   - In each (model class, task type) block, the BEST Task Rec.
%     cell is highlighted in pink and bolded. Nothing else is tinted.
%   - Bold in Recall / Prec. columns marks the best entry per
%     column within each model class (kept from the original table).
% =============================================================

% --- colours (re-use main table palette; \providecolor avoids re-def clash) ---
\providecolor{pinkL}{HTML}{FCE4EC}
\providecolor{blueL}{HTML}{D9EBFB}   % Ground+Act regressions on Process Task Rec.

% best-Task-Rec. cell: pink background + bold value
\providecommand{\hcbest}[1]{\cellcolor{pinkL}\textbf{#1}}
% blue-highlight cell (callout, no bold)
\providecommand{\hcblue}[1]{\cellcolor{blueL}#1}

% =============================================================

\begin{table}[t]
\centering
\small
\setlength{\tabcolsep}{4.5pt}
\renewcommand{\arraystretch}{1.1}
\caption{
Grounding and task-structure recovery diagnostics on the VirtualHome split of FullHome.
\textbf{Recall} / \textbf{Prec.} measure scene-entity coverage, and \textbf{Task Rec.} measures recovery of hidden task conditions.
\tcsq{pinkL}\,\textbf{bold} marks the best Task Rec. within each model class; \tcsq{blueL}\, highlights the Ground+Act process-task regression despite high recall; uncolored \textbf{bold} marks the best Recall / Prec. entry.
}
\label{tab:grounding_diagnostics}
\begin{tabular}{llccc ccc}
\toprule
\multirow{2}{*}{\textbf{Model}} &
\multirow{2}{*}{\textbf{Method}} &
\multicolumn{3}{c}{\textbf{Goal Task}} &
\multicolumn{3}{c}{\textbf{Process Task}} \\
\cmidrule(lr){3-5} \cmidrule(lr){6-8}
& & Recall & Prec. & Task Rec. & Recall & Prec. & Task Rec. \\
\midrule

\multirow{4}{*}{GPT-5}
& Naive        & 80.8          & \textbf{75.5} & 77.9         & 78.0          & \textbf{86.5} & 49.8         \\
& Retrieve+Act & 88.8          & 40.1          & 76.3         & 83.9          & 56.3          & 51.6         \\
& Ground+Act   & \textbf{96.0} & 32.4          & 81.8         & 90.0          & 47.4          & 49.3 \\
& TaskGround   & \textbf{96.0} & 31.7          & \hcbest{91.2}& \textbf{91.7} & 47.5          & \hcbest{67.2}\\
\cmidrule(lr){1-8}

\multirow{4}{*}{Qwen3.5-9B}
& Naive        & 37.4          & \textbf{43.2} & 18.0         & 27.1          & 36.9          & 21.3         \\
& Retrieve+Act & 50.5          & 22.7          & 19.5         & 44.6          & 35.5          & 25.0         \\
& Ground+Act   & \textbf{86.5} & 35.8          & 43.8         & \textbf{80.7} & \textbf{50.3} & \hcblue{15.8} \\
& TaskGround   & 85.6          & 37.3          & \hcbest{79.3}& 80.5          & 49.7          & \hcbest{52.5}\\

\bottomrule
\end{tabular}
\vspace{-0.6em}
\end{table}

\subsection{Grounding Recall Is Necessary but Not Sufficient}
\label{sec:grounding_diagnostics}

Table~\ref{tab:grounding_diagnostics} analyzes whether scene grounding recovers the context needed for task-structure recovery.
\textbf{Retrieve+Act} directly asks the model to select relevant object IDs before action generation, while \textbf{Ground+Act} uses TaskGround's grounded scene slice but still directly generates actions.
\textbf{Task Rec.} measures the fraction of recovered hidden task conditions: goal predicates for goal tasks, and both goal and process conditions for process tasks.
The results show that grounding recall is necessary but not sufficient.
Naive prompting often has higher precision but much lower recall, especially for Qwen3.5-9B, indicating that it focuses on obvious entities while missing task-relevant context.
TaskGround substantially improves recall, but Ground+Act still lags behind TaskGround in Task Rec.
The blue-highlighted Qwen3.5-9B process result illustrates this failure mode: Ground+Act retrieves a much higher-recall scene slice than Retrieve+Act, but obtains lower Task Rec. because the compact model still has to generate skill-level actions directly.
With more relevant entities exposed, it tends to produce longer and more error-prone action sequences, where missing preconditions can terminate execution.
TaskGround resolves this by converting the grounded scene slice into executable task structure before skill-level execution, substantially improving Task Rec.
\section{Related Work}
\subsection{Language Models for Household Embodied Agents}
Large language models have been widely used as high-level decision-making modules for embodied agents.
Prior work has explored LLMs for generating action sequences, programs, or hierarchical plans from natural-language instructions~\citep{llmzeroshot,progprompt,llmplanner,tapa,sayplan,mldt}, grounding high-level decisions in executable skills, affordances, robot policies, or motion-planning feedback~\citep{saycan,codeaspolicies,llm3}, and supporting interactive, uncertainty-aware, or personalized assistance~\citep{innermonologue,knowno,tidybot}.
These works demonstrate the promise of LLMs for embodied decision making, but generally operate on stated instructions, goals, task descriptions, or bounded contexts where the task intent is explicit or can be clarified within the given interface.
More closely related are methods that combine LLM planning with structured retrieval, symbolic intermediates, or graph-based planning: EMBODIED RAG retrieves task-relevant subgraphs for LLM-based planning~\citep{embodiedrag}; UniPlan and DC-SGG use symbolic scene parsing and PDDL-style planning interfaces~\citep{uniplan,dcsgg}; BrainBody-LLM, LookPlanGraph, and SG2 improve embodied planning through feedback, simulator checks, or scene-graph reasoning~\citep{brainbodyllm,lookplangraph,sg2}; and MomaGraph studies task-oriented structured intermediates with learned graph policies~\citep{momagraph}.
TaskGround shares the motivation of using structured intermediates, but differs by targeting the full-scene information condition and by explicitly inferring and completing executable task structure in a training-free, model-agnostic pipeline before skill-level execution.

\subsection{Embodied Household Benchmarks and Evaluation Protocols}
Embodied household benchmarks and simulators provide executable environments for
evaluating household agents.
VirtualHome represents household activities with programmatic action
specifications~\citep{virtualhome}, while BEHAVIOR and BEHAVIOR-1K define
activities through predicate-based initial and goal conditions~\citep{behavior,behavior1k}.
Recent embodied-agent interfaces and planning benchmarks further evaluate
LLM-based agents with structured inputs, action spaces, constraints, or
evaluation functions~\citep{eai,partnr,lotabench}.
Instruction-following and dialogue benchmarks such as ALFRED, ALFWorld, TEACh,
and DialFRED test grounded execution and interactive assistance~\citep{alfred,alfworld,teach,dialfred}.
Beyond these simulator-backed and instruction-following settings, recent
benchmarks have moved toward more user-facing household scenarios, including
personalized memory use~\citep{memento}, vague referring expressions in robot
task planning~\citep{reibench}, and smart-home evaluation with contextual or
personalized instructions~\citep{homebench,personalhomebench}.
These works study important factors beyond clean instruction following, but
usually isolate one factor such as memory, reference ambiguity, device control,
or planner evaluation.
In contrast, \textbf{FullHome} evaluates the full-scene information condition:
agents receive a complete household scene and a situated household request, but
not task-relevant objects, explicit goals, or process constraints.
It measures whether agents can infer executable task structure before execution.

\section{Discussion}

Our results show that full-scene household reasoning is not merely a long-context action-generation problem: agents must identify task-relevant scene context, infer implicit goals and process requirements, and produce executable skill-level actions.
TaskGround separates these difficulties into scene grounding, task-structure inference with completion, and skill-level execution, yielding consistent gains across model families.
Our analyses also show that high-recall grounding alone is insufficient, while oracle scene-slice results indicate that both grounding and task-structure inference remain bottlenecks.
FullHome is a controlled simulator-backed setting with structured scenes and predefined skills; future work should extend this setting to perception noise, partial observability, clarification dialogue, household preferences, safety monitoring, and robust failure recovery.

\section{Conclusion}

We formalized \textbf{full-scene household reasoning}, where agents must act
from a complete household scene and a situated household request rather than a
curated task specification. To tackle this problem, we proposed \textbf{TaskGround}, a training-free \emph{Ground--Infer--Execute}
framework that grounds complete scenes, infers and completes executable task
structure, and compiles it into grounded skill-level actions.
We also introduced \textbf{FullHome}, a simulator-backed evaluation suite for
systematically measuring this setting.
Experiments show that TaskGround consistently improves both proprietary and
open-weight models, enables compact models to approach frontier direct
complete-scene prompting with lower total input-token cost, and identifies
executable task-structure inference as a central bottleneck for practical
household agents.

\clearpage
\bibliographystyle{plainnat}
\bibliography{refs}

% \clearpage
% \input{checklist.tex}

\clearpage
\appendix

\section{Appendix}
\label{app:overview}

\vspace{0.8em}
\noindent{\large\bfseries Table of Contents}
\vspace{0.25em}

\noindent\rule{\linewidth}{0.45pt}
\vspace{0.55em}

\appgroup{Benchmark and Method Details}

\appentrydesc{app:benchmark_details}{Benchmark Details}
{Describes the construction and validation of \textsc{FullHome}, including the evaluation condition, task subsets, construction yield, simulator-in-the-loop construction pipeline, and coverage statistics.}

\appentrydesc{app:taskground_impl}{TaskGround Details}
{Provides implementation details of \textsc{TaskGround}, including the Ground--Infer--Execute decomposition, scene grounding, task-structure inference and completion, skill-level execution, and total input-token cost.}

\appentrydesc{app:prompt_templates}{Prompt Templates}
{Lists the prompt templates used for scene grounding, task-structure inference, ordering-sensitive goal prediction, and direct action sequencing baselines.}

\appgroup{Experiments and Analysis}

\appentrydesc{app:experiment_details}{Experiment Details}
{Provides additional details on evaluated models, serving infrastructure, output parsing and failure handling, VirtualHome and BEHAVIOR execution, and scene grounding diagnostics.}

\appentrydesc{app:qualitative}{Qualitative Examples and Failure Cases}
{Presents representative qualitative examples and recurrent failure modes of direct full-scene action generation, together with how \textsc{TaskGround} mitigates them through grounding, task-structure inference, completion, and skill-level execution.}

\appgroup{Reproducibility and Responsible Use}

\appentrydesc{app:broader_impacts}{Broader Impacts}
{Discusses potential benefits, risks, and mitigation considerations for full-scene household reasoning and household embodied agents.}

\appentrydesc{app:compute_resources}{Compute Resources}
{Reports compute resources used for local open-weight model inference, provider-served model evaluation, simulator execution, and automatic evaluation.}

\appentrydesc{app:asset_licenses}{Asset Licenses}
{Summarizes external assets used in this work, including household simulators, benchmark environments, model providers, open-weight models, and software dependencies.}

\vspace{0.3em}
\noindent\rule{\linewidth}{0.45pt}

\clearpage

\clearpage
\section{Benchmark Details}
\label{app:benchmark_details}

This section provides additional details on the construction and validation of
\textsc{FullHome}, the evaluation suite used to study full-scene household
reasoning.
Each task instance provides a \emph{complete household scene} and a
\emph{situated household request}, and the agent must output a
\emph{grounded skill-level action sequence}.
The hidden task structure, including final-state goals and process constraints
when applicable, is used only for evaluation and is not provided to the agent.

\textsc{FullHome} contains 400 carefully designed and human-validated household
tasks built on executable household simulation environments, including
VirtualHome~\citep{virtualhome} and BEHAVIOR~\citep{behavior}.
The benchmark is constructed through an over-generation and quality-gating
pipeline: candidate tasks are generated from simulator-grounded scene
inventories and household scenarios, and only tasks that pass simulator
validation and human review are retained.

\subsection{Evaluation Condition}

\textsc{FullHome} changes the test-time information condition of existing
simulator-backed household evaluations.
The agent receives only the initial complete household scene \(\mathcal{S}_0\)
and the situated household request \(u\), and must output a grounded skill-level
action sequence \(a_{1:T}\).
It is not given task-relevant object lists, candidate action arguments, explicit
goals, target states, process constraints, or reference action sequences.
Thus, the agent must infer task-relevant scene elements, intended goal
conditions, and process requirements before producing executable grounded
actions.

The benchmark contains four subsets:
\begin{itemize}[leftmargin=1.6em, itemsep=0.25em, topsep=0.25em]
    \item \textbf{VG-200}: 200 VirtualHome goal tasks, evaluated by final-state satisfaction.
    \item \textbf{VP-100}: 100 VirtualHome process tasks, evaluated by both final-state and process-constraint satisfaction.
    \item \textbf{BG-65}: 65 BEHAVIOR goal tasks, used to evaluate cross-environment generalization.
    \item \textbf{BP-35}: 35 BEHAVIOR process tasks, evaluated by both executable goals and process constraints.
\end{itemize}
Here, \textbf{V}/\textbf{B} denote the VirtualHome and BEHAVIOR subsets, and
\textbf{G}/\textbf{P} denote goal and process tasks.

\begin{table}[t]
\centering
\small
\setlength{\tabcolsep}{6pt}
\renewcommand{\arraystretch}{1.1}
\caption{
Construction yield of \textsc{FullHome}.
We over-generate candidate tasks and retain only those that pass simulator
validation, process differentiation when applicable, and human review.
}
\label{tab:construction_yield}
\begin{tabular}{lccc}
\toprule
\textbf{Subset} & \textbf{Candidates} & \textbf{Released} & \textbf{Yield} \\
\midrule
VG-200 & 300 & 200 & 66.7\% \\
VP-100 & 300 & 100 & 33.3\% \\
BG-65  & 120 &  65 & 54.2\% \\
BP-35  & 120 &  35 & 29.2\% \\
\midrule
Total  & 840 & 400 & 47.6\% \\
\bottomrule
\end{tabular}
\vspace{-0.5em}
\end{table}

\subsection{Construction Pipeline}

We construct \textsc{FullHome} through a simulator-in-the-loop pipeline that
over-generates candidate tasks and then filters them through automatic and human
validation.
The goal is to obtain situated household tasks that are executable,
unambiguous, and automatically evaluable, while still requiring agents to infer
what should be done from a complete household scene and a situated request.

\noindent\textbf{Stage 1: Scene inventory extraction.}
We first extract the scene inventory from each simulator environment.
For VirtualHome, we use the evolving-graph representation to obtain rooms,
objects, object states, affordances, and relations.
For BEHAVIOR, we use object-rich multi-room household scenes and parse the
corresponding scene assets and task specifications to obtain rooms, fixed
furniture, sampleable objects, supported predicates, and object affordances.
This produces a structured inventory for each scene, which serves as the basis
for task design and is also the complete household scene provided to evaluated
agents.

\noindent\textbf{Stage 2: Household scenario design.}
We design household scenarios to cover common daily activities and deployment
contexts, such as cleaning, meal preparation, object organization, workspace
setup, guest preparation, appliance use, lighting, rest, and safety routines.
The scenario design is grounded in the extracted scene inventory: candidate
tasks are required to refer only to entities, states, and affordances available
in the corresponding household scene.
This prevents tasks from depending on hallucinated objects or unsupported
actions.

\begin{table}[t]
\centering
\small
\setlength{\tabcolsep}{5pt}
\renewcommand{\arraystretch}{1.1}
\caption{
Scene, room, and theme coverage of \textsc{FullHome}.
For VirtualHome, scenes correspond to functional rooms within the
apartment graph; for BEHAVIOR, scenes correspond to (BDDL activity,
house) base configurations.
Themes are the per-task LLM-generated descriptions retained after the
construction pipeline; each task corresponds to one theme on BEHAVIOR,
while VirtualHome uses a smaller theme pool with multiple tasks per
theme.
}
\label{tab:scene_coverage}
\begin{tabular}{lcccc}
\toprule
\textbf{Subset} & \textbf{\#Tasks} & \textbf{\#Houses} & \textbf{\#Scenes} & \textbf{\#Themes} \\
\midrule
VG-200  & 200 & 1 & 4 rooms        & 98 \\
VP-100  & 100 & 1 & 4 rooms        & 87 \\
BG-65   &  65 & 4 & 10 base scenes & 65 \\
BP-35   &  35 & 3 &  9 base scenes & 34 \\
\midrule
Union   & 400 & \textbf{5} & \textbf{20} & $\sim$\!280 \\
\bottomrule
\end{tabular}
\vspace{-0.5em}
\end{table}

\noindent\textbf{Stage 3: Situated request and hidden task-structure authoring.}
For each scenario, we author a situated household request together with its
hidden task structure.
The request is written from the user's perspective and is designed to be natural
rather than a clean task specification.
It may leave task-relevant objects, target states, or ordering requirements
implicit, but the intended task must be recoverable from the complete scene
context.
We annotate the corresponding hidden task structure
\(\tau=(\mathcal{G},\mathcal{P})\), where \(\mathcal{G}\) contains final-state
goal predicates and \(\mathcal{P}\) contains process constraints when
applicable.
Both goals and process constraints are grounded to entities in the simulator
scene and are never included in the model input.

\noindent\textbf{Stage 4: Ground-truth execution and simulator validation.}
Each candidate task is paired with a ground-truth skill-level execution.
We generate candidate executions using rule-based templates for canonical
household motifs and LLM-assisted drafting for long-tail scenarios.
The resulting action sequence is executed and checked in the corresponding
simulator.
A task is retained only if the execution succeeds and the final state satisfies
all annotated goal predicates.
For process tasks, the execution trajectory must also satisfy the annotated
process constraints.

For VirtualHome, we validate executions with its deterministic evolving-graph
transition system.
For BEHAVIOR, whose underlying simulator can involve continuous physics and
online object instantiation, we instantiate the candidate scene and snapshot the
resulting object configuration, including object states, parent-child relations,
and predicate truth values.
We then run a deterministic graph-style satisfaction check over the snapshot.
This preserves BEHAVIOR task semantics while making evaluation reproducible.

\noindent\textbf{Stage 5: Process differentiation.}
For process tasks, we additionally verify that process constraints are
non-trivial.
We check that the annotated process requirements distinguish valid executions
from alternative executions that may reach the same final state but violate the
required ordering or miss an intermediate condition.
This ensures that process-task success is not reducible to final-state goal
satisfaction alone.

\noindent\textbf{Stage 6: Human review and final filtering.}
All surviving tasks are reviewed for request naturalness, clarity, grounding
consistency, executability, and alignment between the situated request and the
hidden task structure.
We remove tasks with ambiguous intent, unsupported object references, unnatural
phrasing, invalid simulator behavior, or mismatches between the request and the
annotated goals or process constraints.
The final benchmark is sampled after validation to maintain diversity across
environments, rooms, household scenarios, and task types.

Table~\ref{tab:construction_yield} summarizes the construction funnel.
Starting from 840 candidate tasks, the pipeline retains 400 tasks after
simulator validation, process differentiation when applicable, and human review.
The lower yield of process tasks reflects the additional ordering and
intermediate-state requirements imposed by \(\mathcal{P}\).

\subsection{Coverage Statistics}

Table~\ref{tab:scene_coverage} summarizes the scene, room, and theme coverage
of \textsc{FullHome}.
The VirtualHome subsets use a single underlying household environment, but
each task is constructed on a per-task scene perturbation that varies object
placements, states, and clutter, supporting controlled analysis in a known-home
setting.
The BEHAVIOR subsets span four iGibson houses and ten distinct (BDDL activity,
house) base scenes, providing a complementary cross-environment evaluation
setting with realistic object layouts and physics-grounded affordances.

The task themes span a wide range of household contexts, including hygiene,
meal preparation, cleanup, workspace setup, guest preparation, appliance use,
lighting and ambience, rest, and safety routines.
Process tasks additionally annotate ordering invariants drawn from common
household routines, such as cleaning before serving, preparing a surface before
placing objects on it, loading before running an appliance, and closing a
container or appliance before activation.

\clearpage
\section{TaskGround Details}
\label{app:taskground_impl}

This section provides implementation details of \textsc{TaskGround} that are
omitted from the main text.
Given a complete household scene \(\mathcal{S}_0\) and a situated household
request \(u\), \textsc{TaskGround} follows the same three-stage
\emph{Ground--Infer--Execute} decomposition defined in
Sec.~\ref{sec:framework}:
\[
\text{ground: } \mathcal{S}_r = R(\mathcal{S}_0,u), \quad
\text{infer: } \hat{\mathbf{g}} = T(\mathcal{S}_r,u),\ 
\tilde{\tau}=C(\hat{\mathbf{g}}), \quad
\text{execute: } a_{1:T}=E(\tilde{\tau},\mathcal{S}_r).
\]
Here, \(R\) is the scene grounder, \(T\) is the task-structure inference
module, \(C\) is the completion module, and \(E\) is the executor.
The output of \(T\), \(\hat{\mathbf{g}}\), is an initially inferred ordered
goal sequence.
The output of \(C\), \(\tilde{\tau}\), is the completed executable task
structure used by the executor.
The scene grounder \(R\) and task-structure inference module \(T\) each make
one LLM call, while \(C\) and \(E\) are deterministic.
At test time, \textsc{TaskGround} never accesses hidden final-state goals,
hidden process constraints, reference action sequences, task identifiers, or
evaluation outcomes.
The same prompts are used across models, and decoding is deterministic
with temperature \(0\).

\subsection{Ground: Scene Grounder}
\label{app:tg_ground}

The grounder produces a compact task-relevant scene slice
\(\mathcal{S}_r \subseteq \mathcal{S}_0\).
It is designed to reduce the input cost of downstream reasoning while preserving
the entities, states, affordances, and local relations needed for executable
task-structure inference.

\paragraph{Entity inventory.}
The complete scene is first converted into a lightweight entity inventory.
For each room, we list the visible object classes and relevant state summaries,
such as whether an object is open, closed, dirty, clean, on, off, or
receptacle-like.
This inventory is substantially shorter than the full scene graph because it
does not serialize all pairwise object relations.

\paragraph{Node-level selection.}
The LLM receives the situated request \(u\) and the entity inventory extracted
from \(\mathcal{S}_0\).
It outputs a set of potentially task-relevant entities
\(\hat{\mathcal{V}}_r\).
The prompt is recall-oriented: the model is instructed to include explicitly
mentioned entities, implicit tools, destinations, containers, surfaces, and
objects that may be required by the household context.

\paragraph{Scene-slice reconstruction.}
The selected entities are resolved to concrete scene-graph nodes by
deterministic lookup.
Unmatched selections are discarded, which prevents the model from inventing
entity identifiers.
We then expand the selected node set with structural context needed for
execution, including containing rooms, the agent node and state, relevant
containers or surfaces, object states, affordances, and local relations.
If \(\mathcal{V}_r\) is the expanded node set, the grounded slice is the induced
subgraph
\[
    \mathcal{S}_r=(\mathcal{V}_r,\mathcal{E}_r), \qquad
    \mathcal{E}_r=\{(v_i,v_j)\in\mathcal{E}_0 \mid v_i,v_j\in\mathcal{V}_r\}.
\]
Thus, the LLM selects task-relevant nodes, while relations are restored
deterministically from the original scene graph.

\subsection{Infer: Task-Structure Inference and Completion}
\label{app:tg_infer}

Given the grounded scene slice \(\mathcal{S}_r\) and request \(u\), the
inference module \(T\) predicts an ordered goal sequence
\[
    \hat{\mathbf{g}}=(\hat{g}_1,\ldots,\hat{g}_M).
\]
Each goal atom describes a desired state or relation over grounded entities,
such as \texttt{CLEAN(table)}, \texttt{ON(cup, table)},
\texttt{CLOSED(dishwasher)}, or \texttt{INSIDE(cup, dishwasher)}.
This goal-level interface asks the LLM to infer what should be achieved,
rather than to directly generate a long skill-level action sequence.

For VirtualHome, goal atoms are represented using node-state and edge-relation
predicates:
\begin{lstlisting}[style=tgcode]
{
  "goals": [
    {"predicate": "CLEAN",  "args": [<id>]},
    {"predicate": "ON",     "args": [<from_id>, <to_id>]},
    {"predicate": "INSIDE", "args": [<from_id>, <to_id>]}
  ]
}
\end{lstlisting}
Allowed state predicates include \texttt{OPEN}, \texttt{CLOSED},
\texttt{CLEAN}, \texttt{ON}, \texttt{OFF}, \texttt{PLUGGED\_IN}, and
\texttt{PLUGGED\_OUT}, and allowed relation predicates include
\texttt{ON} and \texttt{INSIDE}.
For BEHAVIOR, the same goal-level interface is mapped to BDDL-supported
predicates.
The model is required to use only entities present in \(\mathcal{S}_r\).
Malformed JSON is recovered when possible, but unsupported predicates,
duplicate goals, initially satisfied goals, and references to non-existent
entity identifiers are rejected automatically.

The completion module \(C\) then transforms the initial goal sequence
\(\hat{\mathbf{g}}\) into a completed executable task structure
\[
    \tilde{\tau}=(\tilde{g}_1,\ldots,\tilde{g}_K).
\]
The completion rules encode fixed household priors over goal patterns, such as
closing an appliance before turning it on, closing a container after insertion,
or cleaning a dirty surface before placing an object on it.
These rules are specified before evaluation and do not use hidden goals,
hidden process constraints, reference actions, task identifiers, or evaluation
outcomes.
Thus, \(C\) is a deterministic test-time completion step rather than an oracle
correction.
Its output \(\tilde{\tau}\) is an ordered goal sequence that contains both
final-state goals and process-critical intermediate goals.

\subsection{Execute: Skill-Level Executor}
\label{app:tg_execute}

The executor \(E\) compiles the completed executable task structure
\(\tilde{\tau}\) into a grounded skill-level action sequence \(a_{1:T}\).
It treats each ordered goal atom as a desired scene-state change and realizes it
using the available household skills, such as navigation, manipulation,
cleaning, container operation, and appliance control.
For example, a goal such as \(\texttt{CLEAN(table)}\) may be realized as a
sequence that navigates to a cleaning tool, grasps it, walks to the table, and
wipes the table.

The executor maintains an internal symbolic state after each emitted action so
that later goals are planned against the latest scene state.
A precondition layer inserts generic steps required for executable actions,
such as navigating near an object before manipulation, holding an object before
placing it, opening a closed container before insertion or retrieval, and
ensuring that the agent has a free hand before grasping.
When \(\tilde{\tau}\) contains ordered intermediate goals, the executor
preserves this ordering.
It does not infer new task goals or process requirements; it only realizes the
completed structure produced by the previous stage.

\subsection{Token and Cost Budget}
\label{app:tg_budget}

We measure input-token cost using the \texttt{gpt-4o} BPE tokenizer over the
actual prompts used in our experiments.
The naive baseline serializes the complete household scene in a single LLM call.
\textsc{TaskGround} uses two LLM calls per task: a grounding call over a
lightweight entity inventory and an inference call over the grounded scene
slice.
Table~\ref{tab:tg_budget} reports the total input-token cost summed over all
LLM calls per task.
The same serialized prompts are reused across evaluated models, so the reported
costs reflect the scale of the scene representation rather than model-specific
prompt changes.

\begin{table}[h]
\centering
\small
\setlength{\tabcolsep}{4.5pt}
\renewcommand{\arraystretch}{1.05}
\caption{
Total input-token cost of \textsc{TaskGround} versus the naive full-scene
baseline on \textsc{FullHome}, summed over all LLM calls per task.
Tokens are measured with the \texttt{gpt-4o} BPE tokenizer over the actual
prompts.
}
\label{tab:tg_budget}
\begin{tabular}{llcccc}
\toprule
\multirow{2}{*}{\textbf{Env.}} & \multirow{2}{*}{\textbf{Subset}}
& \textbf{Naive} & \multicolumn{2}{c}{\textbf{\textsc{TaskGround}}} & \multirow{2}{*}{\textbf{Reduction}} \\
\cmidrule(lr){4-5}
& & \textbf{1 call} & \textbf{Ground} & \textbf{Infer} & \\
\midrule
\multirow{2}{*}{VirtualHome}
   & VG-200 & \(\sim\)84{,}300 & \(\sim\)2{,}030 & \(\sim\)2{,}680 & \(\mathbf{17.9\times}\) \\
   & VP-100 & \(\sim\)84{,}300 & \(\sim\)2{,}030 & \(\sim\)2{,}660 & \(\mathbf{18.0\times}\) \\
\midrule
\multirow{2}{*}{BEHAVIOR}
   & BG-65  & \(\sim\)30{,}800 & \(\sim\)2{,}000 & \(\sim\)3{,}300 & \(\mathbf{5.8\times}\) \\
   & BP-35  & \(\sim\)23{,}800 & \(\sim\)2{,}000 & \(\sim\)4{,}100 & \(\mathbf{3.9\times}\) \\
\bottomrule
\end{tabular}
\end{table}

The reduction varies with the scale of the underlying household scene, but
\textsc{TaskGround} consistently lowers total input-token cost by replacing
complete-scene serialization with a lightweight grounding call followed by
task-structure inference over a compact scene slice.

\clearpage
\section{Prompt Templates}
\label{app:prompt_templates}

This appendix provides the prompt templates used in our evaluation to support reproducibility.
Placeholders are shown in \texttt{\{braces\}}.
For all prompting-based stages, models are instructed to use only the entities, states, relations, and action primitives provided in the input.
The prompts do not include hidden final-state goals, process constraints, oracle actions, task identifiers, or evaluation outcomes.

\subsection{Scene Grounding Prompt}
\label{app:prompt_grounding}

The scene grounding prompt receives the situated household request and a room-grouped object catalog derived from the complete household scene.
It asks the model to identify task-relevant object classes and rooms.
The system then resolves selected class names to scene-graph node IDs and applies structural expansion, such as adding relevant rooms, containers, surfaces, power-related objects, and local relational context.

\begin{tcolorbox}[colback=promptgray, colframe=gray!50, title=Scene Grounding Prompt, fonttitle=\bfseries\small, breakable]
\small\ttfamily
You are a scene retriever for an embodied household agent.\\[4pt]
REQUEST:\\
\{request\}\\[4pt]
SCENE OBJECTS, grouped by room. Each line contains: class\_name [count] | states | properties.\\
\{catalog\}\\[4pt]
Task: decompose the request into individual goals or requirements. For each goal, list the (room, object\_class) pairs the robot needs to read or interact with. You do not need to enumerate object IDs or count instances. The system will automatically include all matching objects in that room and add containers, surfaces, rooms, and local context through structural expansion.\\[4pt]
GUIDELINES:\\
- object\_class must be the exact class\_name shown in the scene table.\\
- room must be the exact room name. Use ``any'' only if the object's room cannot be determined from the request.\\
- Include implicit tools even if they are not mentioned by name:\\
\quad wipe or clean a surface -> include rag or towel.\\
\quad wash an object -> include sink and faucet.\\
- Include implicit objects for high-level intents:\\
\quad set up for work -> include computer, keyboard, and mouse if present.\\
- Include destinations for placement requests. For ``place A on/in B'', list B as well.\\
- Be inclusive: missing a relevant object may cause task failure.\\[4pt]
Output only a JSON object with the following format:\\
\{``goals'': [\{``goal'': ``...'', ``selections'': [\{``room'': ``...'', ``objects'': [...]\}]\}]\}
\end{tcolorbox}

\subsection{Task-Structure Inference Prompt}
\label{app:prompt_inference}

The task-structure inference prompt receives the situated household request and the grounded scene slice produced by the scene grounding stage.
It asks the model to infer the executable task structure, including final-state goals and process-relevant ordering requirements, using only entities in the grounded scene slice.

\begin{tcolorbox}[colback=promptblue, colframe=blue!40, title=Task-Structure Inference Prompt, fonttitle=\bfseries\small, breakable]
\small\ttfamily
You are a household robot planner. Given a situated household request and the current grounded scene graph, infer the target goal states and process-relevant requirements needed to complete the task.\\[4pt]
Output two types of final-state goals:\\
1. Node goals: an object should reach a specific state.\\
2. Edge goals: an object should be moved to a specific location or container.\\[4pt]
Possible target states for node goals:\\
ON, OFF, OPEN, CLOSED, CLEAN, PLUGGED\_IN, PLUGGED\_OUT\\[4pt]
Possible relations for edge goals:\\
ON, INSIDE\\[4pt]
Rules:\\
1. Only include changes from the current state.\\
2. Use exact object IDs from the grounded scene graph.\\
3. Include all goals implied by the request and the household context.\\
4. Do not invent objects that are not present in the grounded scene graph.\\
5. If ordering is important, list goals in the order they should be achieved.\\[4pt]
REQUEST:\\
\{request\}\\[4pt]
NODES:\\
\{nodes\}\\[4pt]
EDGES:\\
\{edges\}\\[4pt]
Output only a JSON object with the following format:\\
\{``node\_goals'': [...], ``edge\_goals'': [...]\}
\end{tcolorbox}

\subsection{Ordering Hint for Goal Sequences}
\label{app:prompt_ordering}

For the goal-sequence variant used by TaskGround, the following ordering hint is appended to the task-structure inference prompt.

\begin{tcolorbox}[colback=promptgreen, colframe=green!40, title=Ordering Hint, fonttitle=\bfseries\small, breakable]
\small\ttfamily
IMPORTANT: ORDER MATTERS.\\
List node\_goals and edge\_goals in the order they should be achieved. Each goal will be executed sequentially against the current scene state. For tasks where order is meaningful, such as cleaning before placing, opening a container before putting an object inside, or preparing an object before serving it, put earlier prerequisites first. For order-independent tasks, any consistent order is acceptable.
\end{tcolorbox}

\subsection{Action Sequencing Prompt for Direct-Action Baselines}
\label{app:prompt_action}

The action sequencing prompt is used by direct-action baselines, including Naive and Ground+Act.
In these settings, the model directly outputs a grounded skill-level action sequence instead of first inferring executable task structure.

\begin{tcolorbox}[colback=promptred, colframe=red!40, title=Action Sequencing Prompt, fonttitle=\bfseries\small, breakable]
\small\ttfamily
The task is to guide the robot to take actions from the current scene state based on a situated household request.\\[4pt]
Unlike goal-conditioned settings, the input does not explicitly provide node goals, edge goals, process constraints, or oracle actions. You must infer the intended target objects, target states, and target relations from the request and the provided scene graph, and then output a valid grounded skill-level action sequence.\\[4pt]
Supported actions include:\\
CLOSE, GRAB, OPEN, PUTBACK, PUTIN, SWITCHON, SWITCHOFF, WIPE, WASH, PLUGIN, PLUGOUT, WALK.\\[4pt]
Key rules:\\
- WALK to an object before interacting with it.\\
- GRAB requires a free hand. If both hands are full, release an object first.\\
- WIPE requires holding a cleaning tool when applicable.\\
- SWITCHON requires the object to be OFF and PLUGGED\_IN when applicable.\\
- Use only object IDs that appear in the provided scene graph.\\[4pt]
REQUEST:\\
\{request\}\\[4pt]
NODES:\\
\{nodes\}\\[4pt]
EDGES:\\
\{edges\}\\[4pt]
Output only a JSON list of action commands.
\end{tcolorbox}

\clearpage
\section{Experiment Details}
\label{app:experiment_details}

This section provides additional experimental details for the results reported in the main paper.
Across all experiments, each task provides a complete household scene and a situated household request.
The agent must output a grounded skill-level action sequence; hidden final-state goals, process constraints, reference action sequences, task identifiers, and evaluation outcomes are never provided to the model.
All prompting-based stages use deterministic decoding with \texttt{temperature=0}, and the same prompt templates are used across models.

\subsection{Evaluated Models and Serving Infrastructure}
\label{app:exp_models}

We evaluate both proprietary and open-weight models to cover a range of model capabilities and deployment assumptions.
The proprietary group includes GPT-4o, GPT-4.1, Gemini-2.5-Flash, and GPT-5.
The open-weight group includes DeepSeek-V4-Flash, MiMo-V2-Flash, Gemma-3-12B, and Qwen3.5-9B.
This selection covers frontier proprietary systems as well as compact models that better reflect local-deployment constraints.

Table~\ref{tab:model_zoo} summarizes the serving setup.
Provider-served models are accessed through the corresponding model provider or routing service available at evaluation time.
Locally served open-weight models are deployed with vLLM in bfloat16 precision; no model weights are updated during evaluation.
For GPT-5, we use medium reasoning effort.

\begin{table}[h]
\centering
\small
\caption{Model zoo and serving infrastructure used in our experiments.}
\label{tab:model_zoo}
\begin{tabular}{lll}
\toprule
\textbf{Model} & \textbf{Access / Backend} & \textbf{Serving setup} \\
\midrule
GPT-4o & Azure OpenAI & Provider-served inference \\
GPT-4.1 & Azure OpenAI & Provider-served inference \\
GPT-5 & Azure OpenAI & Provider-served inference, medium reasoning effort \\
Gemini-2.5-Flash & Provider / routing service & Provider-served inference \\
DeepSeek-V4-Flash & Provider / routing service & Hosted open-weight inference \\
MiMo-V2-Flash & Provider / routing service & Hosted open-weight inference \\
Gemma-3-12B & vLLM local serving & 8$\times$ A100-40G or 4$\times$ A100-80G \\
Qwen3.5-9B & vLLM local serving & 8$\times$ A100-40G or 4$\times$ A100-80G \\
\bottomrule
\end{tabular}
\end{table}

\subsection{Output Parsing and Failure Handling}
\label{app:exp_parsing}

All model outputs are parsed automatically.
For direct-action methods, the parser expects a grounded skill-level action sequence with supported action names and grounded entity arguments.
For task-structure inference, the parser expects structured goal atoms over entities in the grounded scene slice.
Manual correction is not applied during evaluation.

If an output contains malformed JSON, the parser attempts to extract the largest valid structured object when possible.
If parsing still fails, the task is recorded as a parse failure.
Outputs that reference nonexistent entities, unsupported predicates, invalid action names, or unsupported action-object combinations are counted as failures unless the issue can be deterministically normalized without changing task semantics.
Transient provider or transport failures are retried up to three times; persistent failures are included in the denominator of all reported metrics.

\subsection{VirtualHome Execution}
\label{app:exp_vh}

For the VirtualHome subsets, we use symbolic execution over evolving household scene graphs.
The simulator updates the scene graph step by step as the predicted grounded skill-level action sequence is executed.
An action may fail if it violates simulator preconditions, such as interacting with an object before walking near it, inserting an object into a closed container, or applying a switch action to a non-switchable object.

For goal tasks, success requires that the final scene state satisfies all annotated goal predicates \(\mathcal{G}\).
For process tasks, success additionally requires that the execution trajectory satisfies the annotated process constraints \(\mathcal{P}\).
All reported scores are computed over the full corresponding split.

\subsection{BEHAVIOR Execution}
\label{app:exp_behavior}

For the BEHAVIOR subsets, tasks are specified using simulator-supported household objects and executable predicate semantics.
To make evaluation comparable across simulator backends, we convert the task state into a discrete graph abstraction whose nodes correspond to household entities and whose predicates encode states and relations such as containment, support, proximity, and object states.

We export and cache task snapshots using servers equipped with 4 NVIDIA RTX A6000 GPUs.
Each snapshot records the scene objects, physical properties, room membership, and predicate-level state needed for prompt construction and evaluation.
Predicted grounded skill-level action sequences are executed against this symbolic task state.
Goal tasks are evaluated by final-state predicate satisfaction, while process tasks additionally check trajectory-level process constraints.
This gives a unified evaluation interface across VirtualHome and BEHAVIOR while preserving each environment's object vocabulary and predicate semantics.

\subsection{Scene Grounding Diagnostics}
\label{app:exp_grounding_metrics}

For grounding diagnostics, we compare the grounded scene slice against oracle task-relevant nodes derived from the hidden task structure.
Let \(\mathcal{V}_r\) denote the node set retained in the grounded slice and \(\mathcal{V}^{\star}\) denote the oracle task-relevant node set.
We compute node recall as
\[
\mathrm{NodeRecall}
=
\frac{|\mathcal{V}_r \cap \mathcal{V}^{\star}|}{|\mathcal{V}^{\star}|}.
\]
This diagnostic is used only for analysis; final task performance is always evaluated by simulator execution.

\clearpage
\providecolor{promptgray}{HTML}{F4F4F4}
\providecolor{caseboxbg}{HTML}{F7F7F7}
\providecolor{caseboxfr}{HTML}{B0B0B0}

\lstdefinestyle{actionseq}{
    basicstyle=\ttfamily\small,
    breaklines=true,
    breakatwhitespace=true,
    columns=fullflexible,
    keepspaces=true,
    showstringspaces=false,
    aboveskip=2pt,
    belowskip=2pt,
}

\section{Qualitative Examples and Failure Cases}
\label{app:qualitative}

We present qualitative examples from the \textsc{FullHome} evaluation set to
illustrate typical failure modes of the naive \emph{Full Scene + Act}
baseline and how \textsc{TaskGround} mitigates them in selected cases.
The examples highlight three recurring issues under complete household
scenes:
(i) \emph{symbol hallucination} under large scene context,
(ii) \emph{affordance violations} where the model applies an operator to an
object lacking the required property, and
(iii) \emph{implicit goal omission} where the model misses goals implied by
the situated household request.
These examples complement the aggregate results in
Section~\ref{sec:experiments} by showing how direct action generation can
fail before or during simulator execution.
In the selected examples, \textsc{TaskGround} mitigates these failures by
grounding the complete household scene into a compact task-relevant scene
slice, inferring executable task structure, completing missing goals or
process-critical subgoals, and delegating action realization to a
rule-based skill executor.

% -----------------------------------------------------------------------------
\subsection{Case 1: Symbol Hallucination Under Large Scene Context}
\label{app:case1_simfail}

\begin{tcolorbox}[colback=caseboxbg,colframe=caseboxfr,
                  title=\textbf{Task VG066: L2 instruction},breakable]
\small
\textit{``Hey, I'm settling in on the office couch for a long, sunny
afternoon of reading and emails. Could you bring the bread from the
dining area over to me and set it right on the couch so I don't have to
get up? The little table by the couch feels grimy, and I want it quiet
in there -- no TV, just a comfy spot.''}
\end{tcolorbox}

\noindent\textbf{Ground-truth goals.}
\texttt{ON(food\_bread.3000, couch.352)},
\texttt{CLEAN(table.355)}, and
\texttt{OFF(television.410)}.

\noindent\textbf{Failure modes under Full Scene + Act.}

\begin{itemize}[leftmargin=*]
\item \textbf{GPT-4o}: \textsc{Sim\_Fail}. The model emits placeholder
strings instead of valid integer IDs:
\begin{lstlisting}[style=actionseq]
1. [WALK]    dining_area    'dining_area_id'    <-- hallucinated ID
2. [FIND]    bread          'bread_id'
3. [GRAB]    bread          'bread_id'
4. [WALK]    office_couch   'office_couch_id'
...
sim_error: invalid literal for int(): 'dining_area_id'
\end{lstlisting}
The complete-scene context contains hundreds of valid IDs; rather than
paying the resolution cost of selecting the right one, the model falls
back to placeholder symbols that the simulator cannot dispatch.

\item \textbf{GPT-4.1}: \textsc{Sim\_Fail}. The model violates the
proximity precondition:
\begin{lstlisting}[style=actionseq]
1. [WALK]      dining_room    201
2. [GRAB]      food_bread     3000          <-- character not close to bread
3. [WALK]      home_office    319
4. [PUTBACK]   food_bread     3000  couch  352
5. [SWITCHOFF] television     410
sim_error: <character>(65) is not close to <food_bread>(3000)
\end{lstlisting}
The bread is inside the freezer, but the model attempts to grab it after
walking only to the room-level node and omits the required \texttt{OPEN}.

\item \textbf{GPT-5}: \textsc{Pass}. The model correctly opens the freezer
first and produces a valid action sequence.
\end{itemize}

\noindent\textbf{Result.}
The naive Full Scene + Act baseline fails for GPT-4o and GPT-4.1, while
GPT-5 succeeds.
Under \textsc{TaskGround}, all three models pass: the scene grounder
reduces ID clutter, and the executor inserts generic precondition actions
such as opening the freezer before grabbing the bread.

% -----------------------------------------------------------------------------
\subsection{Case 2: Affordance Violation on a Non-Switchable Object}
\label{app:case2_affordance}

\begin{tcolorbox}[colback=caseboxbg,colframe=caseboxfr,
                  title=\textbf{Task VG016: L2 instruction},breakable]
\small
\textit{``I'm calling it a night and heading to bed. Do a quick sweep so
the place feels calm and dark with no screens left glowing, and the
bedroom set for sleep. I'm not planning to read or work anymore.''}
\end{tcolorbox}

\noindent\textbf{Ground-truth goals.}
\texttt{OFF(tablelamp.97)}, \texttt{OFF(tablelamp.98)},
\texttt{OFF(light.169)}, \texttt{OFF(light.245)},
\texttt{OFF(television.410)}, and \texttt{OFF(computer.417)}.

\noindent\textbf{Failure mode under Full Scene + Act.}
Both GPT-4o and GPT-4.1 attempt to switch off \texttt{ceilinglamp.96},
which lacks the required \texttt{HAS\_SWITCH} property:
\begin{lstlisting}[style=actionseq]
1. [WALK]      ceilinglamp  96
2. [SWITCHOFF] ceilinglamp  96       <-- no HAS_SWITCH affordance
sim_error: <ceilinglamp>(96) does not have a switch
\end{lstlisting}
Although the full scene graph encodes object affordances, the direct-action
baseline fails to filter candidates by the required property when many
visually or semantically similar objects coexist.
GPT-5 avoids the non-switchable lamp and passes.

\noindent\textbf{Result.}
Under \textsc{TaskGround}, the scene grounder retrieves task-relevant
switchable objects (table lamps, lights, television, computer) and the
executor enforces action preconditions, so \texttt{SWITCHOFF} is never
applied to objects without \texttt{HAS\_SWITCH}.
All selected models pass under \textsc{TaskGround} on this task.

% -----------------------------------------------------------------------------
\subsection{Case 3: Implicit Goal Omission (Schema-Level)}
\label{app:case3_implicit}

\begin{tcolorbox}[colback=caseboxbg,colframe=caseboxfr,
                  title=\textbf{Task VG133: L2 instruction},breakable]
\small
\textit{``Sun's out and I've only got a few minutes for lunch before my
next call. I'm going to throw together a quick sandwich at the kitchen
counter and sip some water at the dining table. Can you get that area
looking ready so I can dig in right away?''}
\end{tcolorbox}

\noindent\textbf{Ground-truth goals.}
\texttt{CLEAN(table.226)},
\texttt{CLEAN(kitchen\_counter.230)}, and
\texttt{CLEAN(cup.2009)}.

\noindent\textbf{Failure modes under Full Scene + Act.}

\begin{itemize}[leftmargin=*]
\item \textbf{GPT-4o}: \textsc{Goal\_Fail}. The model wipes the surfaces
but never washes the cup:
\begin{lstlisting}[style=actionseq]
1. [WALK]  rag 3001
2. [GRAB]  rag 3001
3. [WALK]  kitchen_counter 230
4. [WIPE]  kitchen_counter 230
5. ...     wipes sink and table
           <-- never washes cup.2009
\end{lstlisting}

\item \textbf{GPT-4.1}: \textsc{Goal\_Fail}. The model produces a short
action sequence that cleans surfaces but skips the cup.

\item \textbf{GPT-5}: \textsc{Pass}. The model retrieves
\texttt{cup.2009}, washes it at the faucet, and wipes both surfaces.
\end{itemize}

The phrase \emph{``sip some water''} implies that a clean cup is needed,
even though the cup is not explicitly named.
The direct-action baseline captures the explicit surface-cleaning goals
while missing this implicit, schema-level requirement.

\noindent\textbf{Result.}
\textsc{TaskGround} mitigates this failure through task-structure inference
and completion: the inference stage predicts explicit goals over the
grounded scene slice, and the completion module augments goals implied by
household schemas (e.g.\ ``serve water'' $\Rightarrow$ ``clean cup''),
enabling the executor to wash the cup before completion.

% -----------------------------------------------------------------------------
\subsection{Summary of Failure Modes}

Table~\ref{tab:failure_modes} summarizes recurrent failure modes of the
naive Full Scene + Act baseline and how the stages of \textsc{TaskGround}
mitigate them in the selected examples.

\begin{table}[h]
\centering
\small
\caption{Recurrent failure modes of the naive Full Scene + Act baseline
and how \textsc{TaskGround} mitigates them in the selected examples.}
\label{tab:failure_modes}
\begin{tabular}{p{0.26\linewidth}p{0.34\linewidth}p{0.32\linewidth}}
\toprule
\textbf{Failure mode} & \textbf{Trigger} & \textbf{Mitigated by} \\
\midrule
Symbol hallucination (\textsc{Sim\_Fail})
& Long complete-scene context raises the resolution cost of selecting
valid IDs, so the model falls back to placeholder strings.
& \textbf{Ground}: the compact scene slice reduces ID clutter and
resolution cost. \\
\midrule
Affordance violation (\textsc{Sim\_Fail})
& The model selects a similarly named object without the required
property (e.g.\ \texttt{HAS\_SWITCH}, \texttt{CAN\_OPEN}).
& \textbf{Ground + Execute}: grounding restricts candidates; the executor
enforces affordance preconditions. \\
\midrule
Missing precondition (\textsc{Sim\_Fail})
& The model forgets generic prerequisite steps (e.g.\ \texttt{OPEN}
before \texttt{GRAB} from a container).
& \textbf{Execute}: the rule-based executor inserts the missing
precondition actions. \\
\midrule
Implicit goal omission (\textsc{Goal\_Fail})
& The model extracts explicit goals but misses goals implied by household
schemas attached to the request.
& \textbf{Infer + Completion}: schema-level completion adds missing goals
and process-critical subgoals. \\
\bottomrule
\end{tabular}
\end{table}

\clearpage
\section{Broader Impacts}
\label{app:broader_impacts}

This work studies full-scene household reasoning for embodied agents, with the goal of evaluating and improving an agent's ability to infer executable task structure from complete household scenes and situated household requests.
A potential positive impact is that such evaluation can help develop household agents that better understand context-dependent user needs, avoid brittle execution from underspecified instructions, and operate more reliably in everyday indoor environments.
By emphasizing compact open-weight models and structured grounding, the work also supports privacy-conscious and resource-efficient deployment settings, where detailed household states need not always be sent to frontier cloud-hosted proprietary models.

At the same time, household embodied agents raise important societal and safety concerns.
Complete household scenes may contain sensitive information about personal spaces, routines, object usage, accessibility needs, or other private aspects of daily life.
If such information is collected, stored, or transmitted without appropriate safeguards, it could create privacy risks.
In addition, incorrect task-structure inference may lead to wrong or unsafe actions, especially when user requests are ambiguous, when the scene representation is incomplete or noisy, or when the robot acts in environments involving fragile objects, appliances, pets, children, or older adults.
There is also a risk that household reasoning systems could be misused for intrusive monitoring or surveillance if deployed without proper consent and access control.

Our work does not deploy robots in real homes and does not claim to solve these issues.
FullHome is intended as a controlled benchmark for studying task-structure inference under simulator-backed evaluation.
Future deployment of similar systems should include privacy-preserving local inference where possible, clear user consent, secure handling of household scene data, uncertainty-aware planning, clarification dialogue for ambiguous requests, safety constraints, and human override mechanisms.
We hope that explicitly evaluating full-scene reasoning and process constraints can help identify failures before such systems are considered for real-world household use.

\clearpage
\section{Compute Resources}
\label{app:compute_resources}

All experiments in this work are conducted in an inference-only setting.
TaskGround is a training-free framework and does not require model fine-tuning or gradient-based optimization.
No model weights are updated during our experiments, and no additional task-specific training data is used.
Therefore, the reported compute resources correspond only to model inference, simulator execution, and automatic evaluation.

For open-weight models, we run inference locally using bfloat16 precision.
In particular, Qwen3.5-9B is evaluated on GPU servers equipped with either 8 NVIDIA A100 GPUs with 40GB memory each or 4 NVIDIA A100 GPUs with 80GB memory each, depending on hardware availability.
These resources are used for model inference over FullHome tasks, including scene grounding, task-structure inference, task-structure completion, and grounded skill-level action generation when applicable.

Simulator execution and automatic evaluation are performed separately from language-model inference.
For tasks requiring interactive household simulation, we use an iGibson-based simulator~\citep{igibson} and run simulator execution on servers equipped with 4 NVIDIA RTX A6000 GPUs.
These resources are used to execute predicted grounded skill-level action sequences, update environment states, and verify final-state goals and process constraints.

For proprietary models, including GPT-family and Gemini-family models, we use API-based inference through the corresponding model providers or API routing services available at evaluation time.
These models are not run on local GPUs, and their compute resources are managed by the external API providers.
For all models, we use the same benchmark inputs and evaluation protocol: each agent receives the initial complete household scene and the situated household request, and outputs a grounded skill-level action sequence for simulator-based execution and evaluation.

\clearpage
\section{Asset Licenses}
\label{app:asset_licenses}

We use existing simulators, benchmarks, model APIs, and open-weight models only for research evaluation.
We credit the original creators of these assets through citations in the main paper and appendix, and follow the corresponding licenses, model licenses, and terms of use.
FullHome does not redistribute the original simulator assets or proprietary model weights.
When our benchmark metadata or code is released, we will include copyright notices, references to the original assets, and license information for all derived files where applicable.

\noindent\textbf{Household simulation environments.}
FullHome builds on existing household simulation environments and executable task interfaces.
VirtualHome is used as a household activity simulation environment and is credited to its original authors~\citep{virtualhome}.
The public VirtualHome repository is released under a permissive open-source license, and its Unity-related assets are accompanied by the license terms provided by the original project.
BEHAVIOR is used as a household activity benchmark and executable task environment, and is credited to its original authors~\citep{behavior}.
The public BEHAVIOR code repository is released under the MIT License, and we follow the license and usage conditions of the benchmark assets.
For interactive simulation, we use iGibson-based environments~\citep{igibson}.
The iGibson code is released under an open-source license, while the associated scene and asset datasets are governed by the corresponding iGibson dataset license agreement.
We use these assets only for simulator execution and automatic evaluation.

\noindent\textbf{Language models.}
For open-weight model evaluation, we use Qwen3.5-9B~\citep{qwen35} under its Apache License 2.0.
For API-based proprietary or hosted models, including GPT-family, Gemini-family, and DeepSeek-family models, we access the models through the corresponding official providers or API routing services available at evaluation time.
Their use is governed by the relevant service terms, model terms, and usage policies of the corresponding providers.
We do not redistribute proprietary model weights or API outputs as standalone model assets.

\noindent\textbf{Software and implementation dependencies.}
Our implementation relies on standard open-source software packages for model inference, simulator control, data processing, and evaluation.
We follow the licenses of these dependencies and will provide a dependency list with version information in the released code.
All released benchmark annotations and task metadata will be distributed with a license compatible with the licenses and terms of the underlying simulators and assets.

\end{document}